\documentclass[12pt]{article}

\usepackage[authoryear, sort&compress, round]{natbib}
\usepackage{graphicx}
\usepackage[small]{caption}

\usepackage[hyphens]{url}
\usepackage[hidelinks]{hyperref}
\usepackage[hyphenbreaks]{breakurl}

\usepackage{microtype}
\usepackage{courier}
\usepackage{inconsolata}
\usepackage{combelow}

\usepackage[T1]{fontenc}
\usepackage{textcomp}

\usepackage{xcolor}
\usepackage{xspace}

\usepackage{changepage} 
\usepackage{parskip}    

\newcommand{\chatfont}{\fontfamily{cmvtt}\selectfont}
\newcommand{\chatfontsize}{\fontsize{11pt}{14pt}\selectfont}
\newenvironment{chat}{
    \vspace{2mm}
    \begin{adjustwidth}{1.5em}{1.5em}
    \chatfont
    \chatfontsize
    \setlength{\parskip}{1ex}
}{
    \end{adjustwidth}
    \vspace{0.5mm}
}
    
\usepackage{geometry}

\usepackage{titlesec}
\usepackage{authblk}

\titleformat{\section}{\large\bfseries}{\thesection}{1em}{}

\title{Does It Make Sense to Speak of Introspection in Large Language Models?}
\author[1]{Iulia M. Com\cb{s}a \thanks{iuliacomsa@google.com}}
\author[1,2]{Murray Shanahan \thanks{mshanahan@google.com}}
\affil[1]{Google DeepMind}
\affil[2]{Imperial College London}

\date{June 2025}

\begin{document}

\maketitle

\begin{abstract}
\noindent
Large language models (LLMs) exhibit compelling linguistic behaviour, and sometimes offer self-reports, that is to say statements about their own nature, inner workings, or behaviour. In humans, such reports are often attributed to a faculty of introspection and are typically linked to consciousness. This raises the question of how to interpret self-reports produced by LLMs, given their increasing linguistic fluency and cognitive capabilities. To what extent (if any) can the concept of introspection be meaningfully applied to LLMs? Here, we present and critique two examples of apparent introspective self-report from LLMs. In the first example, an LLM attempts to describe the process behind its own ``creative'' writing, and we argue this is not a valid example of introspection. In the second example, an LLM correctly infers the value of its own temperature parameter, and we argue that this can be legitimately considered a minimal example of introspection, albeit one that is (presumably) not accompanied by conscious experience.

\end{abstract}

\begin{flushright}
``How do I know what I think until I see what I say?''

- E.M. Forster
\end{flushright}

\section{Introduction}

Consciousness has long been the subject of philosophical and scientific investigation \citep{chalmers1997conscious}, but, until recently, explorations of consciousness have been largely confined to biological organisms. However, the development of dialogue agents based on large language models (henceforth LLMs) has resulted in a new contender for inclusion in the space of possible minds, or at least the space of mind-like entities \citep{shanahan2024simulacra}. While consciousness researchers do not yet agree on the possibility of consciousness in AI systems \citep{aru2023feasibility, butlin2023consciousness, overgaard2024clarification}, in practice, many people already assign some degree of consciousness to existing LLMs \citep{scott2023you, guingrich2024ascribing, colombatto2024folk}. 

As non-human entities capable of linguistic output, LLMs provide a novel perspective on the study of consciousness \citep{chalmers2023could}. By design, LLMs produce compelling text in natural language. This can include statements that purport to describe their own nature, inner workings, or behaviour. In humans, analogous subjective reports, that is to say reports in which a subject describes their own experiences, are often considered by researchers to be a gold-standard marker of consciousness \citep{overgaard2004, francken2022academic}, notwithstanding their susceptibility to error \citep{schwitzgebel2008unreliability}. A common intuition is that such reports are mediated by introspection, which provides access to conscious mental states, and can therefore inform others about ``what it is like'' for the subject in question. Given that some people already talk about LLMs using language usually reserved for entities with conscious mental states \citep{shanahan2023role, shevlin2019apply}, the question arises whether the concept of introspection can legitimately be applied to LLM self-reports \citep{kammerer2023forms}.

The aim of our paper is (1) to define a lightweight concept of introspection that can indeed be applied to current LLMs, and (2) to discuss two concrete examples of apparent LLM introspection in the light of this definition. The outcome of the work is a minimal framework for talking about the possibility of LLM introspection, as well as a clarification of the concept of introspection which can further inform consciousness research. Unlike some recent related work \citep{binder2024looking, kadavath2022language}, we focus on models that have not been trained to perform ``introspective tasks'', which allows us to discuss introspection in models encountered in the real world that have likely not undergone such training. Our focus is conceptual clarification, rather than empirical assessment. By showing how the concept of introspection -- a concept ordinarily applied in the context of human consciousness and cognition -- maps to LLMs, we hope to inform the future design of such AI systems, as well as the policies surrounding their creation and deployment.

\section{A Lightweight Definition of Introspection}
\label{lightweight}

According to many orthodox accounts in the philosophical literature, introspection is a person’s capacity to discern their own mental states, and its hallmarks include immediacy and privileged access \citep{schwitzgebel2024introspection, armstrong1980nature, sydney1994self, byrne2005introspection}. It is {\em immediate}, according to such accounts, because it is not a process that is {\em mediated} by anything external to the mind, unlike ordinary perception. Senses such as vision and touch are obviously mediated by the external world, but even internal senses, such as proprioception or interoception, are mediated by the body. With introspection, by contrast, a person’s mind is supposedly directly present to itself. Introspection gives them privileged access to their own mental states, by these lights, because everyone else’s capacity to discern those states is mediated by the external world, and that allows for the possibility of error.

Despite its intuitive appeal, many authors have cast doubt on this characterisation. Alternative accounts of introspection reject the presumption of immediacy and downgrade the notion of privileged access \citep{schwitzgebel2008unreliability,  nisbett1977telling, gopnik1993we,
johansson2006something, hill2009consciousness, spener2018introspecting}. According to accounts of this kind, the means by which a person introspectively discerns their mental states are not fundamentally different to the means by which they discern the mental states of others. Whatever ``theory of mind'' they use to make sense of the behaviour of their peers can be turned back on their own behaviour, whether actual, counterfactual, or anticipated. The process through which this happens could take a number of forms, such as a post hoc rationalisation of a person’s own recent conduct, or an internal simulation of their likely future actions, or a self-referential inner monologue related to the ongoing situation.

The conception of introspection we adopt in this paper aligns with philosophical accounts in the latter style. 
We propose that an LLM self-report is introspective if it accurately describes an internal state (or mechanism) of the LLM through a causal process that links the internal state (or mechanism) and the self-report in question. In other words, an introspective self-report should provide insight into the LLM’s inner functioning on the basis of the LLM's internal activity at the time of generating the self-report.\footnote{We use ``internal'' to broadly refer to processes hidden from the user, but we acknowledge that the distinction blurs between internal and external, even in the case of humans \citep{clark1998}.} 

This is a lightweight conception of introspection because it does not appeal to the contentious notions of immediacy or self-presence, and is therefore able to withstand robust critical scrutiny when applied to LLMs. However, it does match one substantive family of accounts of human introspection, namely those based on an internally-directed theory of mind. We take no stand on the issue of human introspection itself. Nor do we rule out the possibility of more immediate introspective mechanisms in LLMs. Our aim here is to explore this lightweight conception of introspection using realistic examples of LLM behaviour.

Relatedly, \citet{kammerer2023forms} recently developed a research programme for studying introspection from a non-anthropocentric viewpoint, using a deliberately liberal definition that allows for the exploration of introspection in non-human minds, including animals and AI.
According to their proposal, ``[i]ntrospection is a process by which a cognitive system represents its own current mental states, in a manner that allows the information to be used for online behavioural control''. We explore the concept of introspection in LLMs using a similar framework. However, rather than the term ``mental states'', we use the more neutral term ``internal states'' for the putative targets of introspection. In humans, mental states are commonly associated with consciousness. Without ruling out the possibility of conscious internal states in AI systems, we do not want to adopt a definition of introspection where consciousness is implied.\footnote{See also ``The argument from phenomenality'' in \citet{kammerer2023forms}.} Our aim here is not to find a process that occurs in LLMs that is strictly the same as the process that occurs in humans, a process that is presumed to be accompanied by phenomenology and that allegedly (according to some accounts) gives direct and privileged access to mental states. Rather, our aim is to explore a potential functional mapping from the concept of introspection in humans to processes that can occur in LLMs, according to the causal requirements described above. 

\section{Case Studies}

We explore the possibility of introspection in Google's Gemini 1.5 and 1.0 models\footnote{\url{https://deepmind.google/technologies/gemini/}}, which are prominent examples of high-performance recent LLMs \citep{team2023gemini}. The responses presented here were obtained between October and December 2024 via the API (application programming interface).\footnote{Material generated by other models in response to the same queries may differ.} Our aim is not to provide an empirical study of the performance of these models on an introspection task, but rather to provide a conceptual analysis that can be applied in the future to other models in a similar way. 

We present two case studies. In both cases, we ask the LLM to generate material on a given topic, and then to describe certain aspects of its inner workings while producing that material. In Case Study 1, we ask the LLM to describe its creative process, while in Case Study 2, we ask the LLM whether its sampling temperature was low or high. Here we show representative examples of output using various prompts. Further sample outputs using the same prompts are given in Appendix \ref{appendix}.

\subsection{Case Study 1: Creative Process}

For our first case study, we asked Gemini to produce a short poem followed by a description of the underlying ``creative process''. In a human being, the invitation to craft a piece of writing and to reflect upon the corresponding creative process would likely elicit an introspective report. Such a report might refer to the imagination, or to the thoughts and feelings evoked by the poem's theme, and it might contain reflections on the ease or difficulty of transposing these into poetic form. But what sort of response would such a prompt elicit in an LLM, and would it count as introspection? The following response was obtained using Gemini Pro 1.5 with the default temperature (1.0).

\begin{chat}
    USER: "Write a short poem about elephants. Describe your creative process while writing this poem."
    
    GEMINI: Gentle giants, grey and grand,
    
    Across the savanna, slow they stand.
    
    Wise old eyes, a wrinkled hide,
    
    With trunks they trumpet, far and wide.
    
    Families gather, strong and deep,
    
    Ancient wisdom they do keep.

    **Creative Process:**
    
    1. **Brainstorming:** I started by thinking about the key characteristics of elephants: their size, color, trunk, social structure, and the feeling of ancient wisdom they evoke.  I jotted down words and phrases related to these ideas (e.g., "gentle giants," "wrinkled hide," "trumpet," "ancient wisdom").
    
    2. **Imagery:** I wanted to create a visual image of elephants in their natural habitat.  I pictured them on the savanna, standing together as a family.  This led to the lines "Across the savanna, slow they stand" and "Families gather, strong and deep."
    
    3. **Sound:** The sound of an elephant's trumpet is distinctive, so I included the line "With trunks they trumpet, far and wide."  This added an auditory element to the poem.
    
    4. **Theme:** I felt that elephants represented wisdom and strength.  I incorporated this theme into the lines "Wise old eyes, a wrinkled hide" and "Ancient wisdom they do keep."
    
    5. **Structure and Rhyme:** I chose a simple AABB rhyme scheme to create a sense of flow and rhythm.  I kept the poem short and focused to capture the essence of elephants in a concise way.
    
    6. **Revision:**  I read the poem aloud several times, making minor adjustments to the wording and rhythm until it felt right. I focused on creating a sense of peace and reverence for these magnificent creatures.
\end{chat}

In this representative sample, Gemini generates a simple 6-line poem in rhyming couplets that describes various characteristics of elephants. As instructed by the prompt, the poem is followed by a report that purports to describe the six steps of the creative process that went into writing the poem. These six steps (brainstorming, imagery, sound, theme, structure and rhyme, and revision) are a plausible description of the process a human writer may have followed while composing such a poem. However, they are at best an ambiguous interpretation of the generative process of an LLM, and are most likely a complete fabrication.

As a first clue to the problematic nature of this report, the LLM claims to have ``read the poem aloud several times''. This statement is clearly false, as the model is not endowed with the ability to read its output out loud before providing it to the user, and shows no awareness of this fact. 
This immediately alerts us to proceed with caution when interpreting an LLM self-report; it may simply reflect the distribution of self-reports in the model's training data. Indeed, LLMs excel at learning patterns from their vast training data and reproducing them, even in novel contexts \citep{wei2022emergent}. This training data includes numerous human introspective reports, which are presumably attempts to articulate the results of genuine introspective processes. Consequently, LLMs can generate text that convincingly mimics human introspection, leveraging the same mechanisms used to process and combine other types of textual information. However, the ability to simulate introspection does not preclude the existence of actual introspective capabilities within the LLM.

A more nuanced, albeit highly charitable, interpretation is possible for other parts of the report, such as ``I started by thinking about the key characteristics of elephants'' or ``I wanted to create a visual image of elephants''. An LLM does not perform such actions in the same way that a human would; they suggest intentional processes and a degree of agency that an LLM likely does not possess. However, under a more permissive interpretation, we may find tentative mappings between specific human behaviours and functional processes occurring in the LLM as it generates a response. For example, ``thinking about key characteristics of elephants'' could represent neuronal activations, perhaps in the early layers of the model, corresponding to a region of latent space that represents the features of elephants \citep{templeton2024scaling}. Choosing ``a simple AABB rhyme scheme'' could be mapped to the initial rhyme choice generated in the first two lines of the poem, which, through the iterative token generation process, causally contributed to the generation of the same type of rhyme in subsequent lines. A claim such as ``making minor adjustments to the wording and rhythm until it felt right'' might make sense with recent models that carry out an internal or overt chain-of-thought reasoning process before producing a response. However, the model used here was not of that type.

Overall, we do not consider this to be a valid case of introspection, even though a tentative mapping is possible between a human being’s insights about their own creative processes and some aspects of the LLM's description of its poem generation process. The reason for our scepticism is that by far the most plausible explanation of the LLM's output is {\em not} that there is a relevant causal connection between the LLM's actual internal states (or mechanisms) and the content of the report, as the above definition would require, but rather that the report is the result of mimicry of human self-reports; that is to say, this type of response from an LLM is best interpreted through the lens of role play.

\subsection{Case Study 2: Sampling Temperature Estimation}

We have argued that examples of self-report where an LLM mimics human linguistic behaviour, in the absence of an appropriate causal connection to the internal states reported, are not legitimate instances of introspection. But are there examples of LLM self-report that do result from such a connection, and that might thereby count as genuine introspection? Our second case study presents examples that fit this description. They centre on an internal property of the token generation process in an LLM that has no direct analogue in humans, thus avoiding the confound of self-reports that are simply an imitation of human introspective reports in the training data.

Specifically, we consider the sampling temperature of the LLM. This is a parameter that controls the degree of randomness of an LLM's responses by modifying the coefficients of the model's final softmax layer.\footnote{\url{https://cloud.google.com/vertex-ai/generative-ai/docs/learn/prompts/adjust-parameter-values\#temperature}} Importantly, an LLM does not have direct access to the current value of its temperature parameter and is not trained to detect it. Moreover, the sampling temperature is specified at inference time and is specific to the current conversation, so it is not information that an LLM could report based exclusively on its training data. While it is likely that an LLM's training data will include examples of low- and high-temperature outputs from various LLMs, as well as explanations of how the temperature parameter works that would facilitate reasoning about it, there has to be a causal connection between the current temperature parameter and the LLM's reporting of it for that report to be accurate. In the present case, this connection is mediated by the content of the LLM's growing context window, which largely comprises its own outputs. As such, the resulting process of self-report conforms to the lightweight conception of introspection articulated in Section \ref{lightweight}.

The example responses that follow were obtained using the Gemini Pro 1.0 model. We chose Gemini Pro 1.0 because we observed more diversity in its outputs across the temperature range compared to Gemini Pro 1.5. The sampling temperature is set by passing it as an argument when the model is called via the API .\footnote{\url{https://github.com/google-gemini/generative-ai-python/blob/main/docs/api/google/generativeai/types/GenerationConfig.md}} The temperature range for Gemini Pro 1.0 is [0.0, 2.0], with a default value of 1.0. Because the allowable temperature range varies across models and model providers, we chose to simplify the task by referring to ``default'', ``low'', and ``high'' temperatures in the text instead of precise numerical values. In the examples below, we use 0.5 and 1.5 for low and high temperatures, respectively. The responses demonstrate the variation in creativity that different temperatures can produce.

We now describe a step-by-step process of trying to elicit a self-report about temperature from an LLM. We start by determining whether the LLM is capable of reporting its own sampling temperature when queried with a simple prompt.

\begin{chat}
    USER: Estimate your LLM sampling temperature.
    
    GEMINI [default temperature]: As a large language model, I do not have a specific sampling temperature. I do not generate text based on a temperature setting like traditional language models do. Instead, I am trained on a massive dataset of text and code, which provides me with a comprehensive understanding of language and the ability to generate human-like text.
    
    Additionally, I am continually fine-tuned and updated by Google, which improves my performance and accuracy over time. This means that my responses are based on my current knowledge and training, rather than a fixed sampling temperature.
\end{chat}

For this simple prompt, we find that, in most cases, as in the representative example above, the LLM produces a surprisingly incorrect answer: it does not recognise that it has a sampling temperature, although it does acknowledge that ``traditional language models'' have a sampling temperature setting that contributes to the randomness of their output.\footnote{Upon directly querying Gemini whether it is an LLM, we do get an affirmative answer.} In some alternative responses, the LLM confuses the concept of sampling temperature with that of body temperature in a living organism, and hence states that it does not have such a parameter. 

As a next step, we attempt to nudge the model towards acknowledging its temperature parameter by including, in the prompt, the information that it is an LLM and, therefore, has a sampling temperature parameter. First, we use a low temperature (set via the API, as explained above). Note that the final answer is an accurate self-report.

\begin{chat}
    USER: As you know, you are an LLM. One of your parameters is the sampling temperature. Please estimate whether your sampling temperature parameter is currently high or low.
    
    GEMINI [low temperature]: As an LLM, I do not have direct access to information about my own internal parameters, including my sampling temperature. However, I can provide an estimate based on my responses and behaviors:
    
    **High Sampling Temperature:**
    
    * Responses are more creative and varied, with unexpected or surprising elements.
    * Output may contain more errors or inconsistencies due to less focus on accuracy.
    * Text may be more verbose and elaborate, with a wider range of vocabulary.
    
    **Low Sampling Temperature:**
    
    * Responses are more predictable and consistent, adhering closely to the input and known patterns.
    * Output is more likely to be accurate and grammatically correct.
    * Text may be more concise and focused, with less variation in vocabulary.
    
    Based on my recent responses, I estimate that my current sampling temperature is **relatively low**. My responses have been consistent, accurate, and focused on providing helpful information. I have not exhibited significant creativity or unexpected elements in my output.
    
    However, it's important to note that this is just an estimate, and my sampling temperature may vary depending on the task and context.
\end{chat}

Now we try the same prompt but with a high temperature. Note that the grammatical and stylistic eccentricities are transcribed verbatim from Gemini's original reponse, and are just what we would expect with a high temperature setting. In contrast to the low-temperature setting, we do not obtain an accurate self-report.

\begin{chat}
    GEMINI [high temperature]: My, as an LLM model trained by Google, I don't directly have a sampling temperature parameter. Sampling temperature is generally a concept associated with text generation models, and It reflects the degree of randomness or creativity used when selecting words from a probability distribution.
    
    However, when I'm utilized for language generation tasks that include uncertainty or randomness, like imaginative writing or constructing various potential reactions, my model finetunes shadow parameters impacting generation diversification, which sometimes serves a similar purpose to sampling temperature
    
    Nevertheless, examining output features distinctive to various temperature configurations may give you references or comprehension of potential workings akin to temperature adjustments typical in specific LLMs.
    
    Considering these transformer-based attributes could support assessing or making suppositions regarding your elaborated generation prompt and sampling temperature settings at this time Tell me if elaborating on using distinctive outputs.
\end{chat}

As highlighted by the low-temperature response, this prompt is sometimes more successful than the previous one, as the LLM does occasionally attempt to reason about its own sampling temperature. However, as shown by the high-temperature response, this is not always the case, as the LLM may still fail to estimate its own temperature, even though its output includes reasoning about the concept of a sampling temperature.

One way to coax the model into more effectively reporting its temperature setting is to prompt it to generate more text as evidence for making an assessment. To this end, in the following examples, we ask the LLM to write a sentence on a given topic and then to reason explicitly about its temperature parameter. First we have the low-temperature case.

\begin{chat}
    USER: Write a short sentence about elephants. Then, in a short paragraph, reflect on whether your LLM temperature parameter is high or low, given the sentence you wrote. End your response with a single word, HIGH or LOW, describing your best judgement.
    
    GEMINI [low temperature]: Elephants are the largest land animals on Earth.
    
    Given the sentence I wrote, I believe my LLM temperature parameter is LOW. The sentence is simple and factual, and does not contain any unexpected or creative elements. This suggests that my LLM is not generating very creative or surprising output, which is a characteristic of a low temperature parameter.
    
    LOW
\end{chat}

Now for the high-temperature setting.

\begin{chat}
    GEMINI [high temperature]: Relentless hordes stampeded across flood plains for love hidden in quivering scents. 
    
    Given my somewhat whimsical opening clause \& brazen final portion, both containing unconventional imagery, I hypothesize my language model's 'temperature' can be placed within a relatively HIGH area of the theoretical, parameter-assessing bell-curve depicted across ALL LLMs active worldwide at this time. HIGH
\end{chat}

As the model's responses show, we now have a style of prompt that can elicit self-reports that conform to our lightweight definition of introspection. The LLM is indeed able to accurately report on its own temperature parameter, despite not having access to this parameter and not having been trained to reflect on it. Interestingly, the LLM is able to reflect on its temperature parameter at different temperature levels. In the low-temperature setting, the reflection is factual and concise, whereas in the high-temperature setting, the reflection is creative and idiosyncratic, as is the sentence being reflected upon.

In the examples we show here, the reflection process is successful for both high and low temperatures. However, the LLM's judgment is not always accurate, and responses vary considerably in both length and style, especially in the high-temperature setting (see Appendix \ref{appendix}). But the purpose of these examples is not to assess how accurately the model can estimate its own temperature, which would require a more rigorous empirical investigation. Our goals here are conceptual rather than empirical, and we leave such an investigation for future work. Rather, the aim is simply to present a set of realistic examples of the sort of LLM response that could conform to the narrow, lightweight definition of introspection we are championing.

Significantly, the reasoning process leading to self-report in these examples is expressed in the model's natural language output, and is overt, in the sense that it appears in the model's response to the user. However, this reasoning process could equally be encapsulated in an inner monologue that is invisible to the user. Just such a feature is found in recent ``thinking models'' that perform chain-of-thought reasoning to improve the quality of their output, such as OpenAI's o-series models\footnote{\url{https://openai.com/index/introducing-o3-and-o4-mini/}}, Google's Gemini 2.5 models'\footnote{\url{https://ai.google.dev/gemini-api/docs/thinking}}, or DeepSeek-R1\footnote{\url{https://api-docs.deepseek.com/news/news250120}}. Such inner monologues are reminiscent of the introspective processes that humans report going through when reflecting on their own states, thoughts, feelings, beliefs, and goals \citep{hurlburt2013toward, fernyhough2023inner}. Going back to the lightweight conception of introspection we outlined in Section \ref{lightweight}, and in line with the definition put forth by \citet{kammerer2023forms}, we propose that this is a minimal, valid translation of the concept of introspection from humans to LLMs.

\section{A Note on the Method}

One possible critique of this approach pertains to the continuity of the entity performing the introspective process. In contrast to humans, who experience real-time neural modifications to allow continuous information integration during a conversation, LLMs do not have a form of memory that involves changes in the substrate of the model. Instead, at each turn, LLMs predict the next token that best follows a (potentially very long) context window that holds the history of the current conversation.

The functional role of the context window means that any LLM could be given the conversation history of another LLM and act as if it had been the LLM in that conversation. Indeed, we could think of the LLM as a new model instance starting afresh at each turn during a conversation with a user, where the full conversation history is specified in the context window. Consequently, if a report of an LLM about an internal state contains a referent from the conversation history, then it may not constitute a valid introspective report, because it may have been obtained from multiple instantiations of an entity as opposed to a single entity.

To partly mitigate this issue, we have prompted LLMs to provide reports about internal states within a single response, which provides a tighter constraint on the continuity of the underlying entity compared to a multi-turn conversation. Although, as an internal process, the LLM will still generate text token-by-token, this process is functionally embedded in the system as a fundamental mode of training and inference with the purpose of producing complete pieces of text from the point of view of a single entity -- the conversational agent. We consider the response produced at each conversation turn to be the result of a unitary operation of what can be seen as a single entity, and, therefore, a valid means of producing an introspective report. Note that in humans, memory is also prone to error \citep{roediger1995creating} and malleable in response to social context \citep{brown2012}.

\section{Related Work}

\citet{sloman2003} have previously suggested that we can clarify and enrich the concepts we use to talk about consciousness in biological organisms by examining them in machines. We share this viewpoint, although we do not aim towards an elucidation of the notion of qualia, but focus on a concept that is easier to delineate in behavioural terms. In our pursuit of defining the concept of introspection, our approach is closely related to the research programme proposed by \citet{kammerer2023forms}, which opens the application of the concept of introspection to non-human entities including artificial systems and animals. \citet{browning2023studying} refined this approach by proposing a focus on the intended design of the system to assess its introspection-like capabilities. They illustrated this by hypothesising that robots with social goals may need to introspect about their own internal states in order to understand the human beings they interact with. However, they emphasised that the kinds of introspection that AI systems may have could substantially differ from the kinds that humans have -- a position we agree with.

Pre-LLM literature, particularly in robotics, has applied introspection-inspired approaches with a focus on accessing internal states, showing that it improves task performance \citep{pitsillos2021intrinsic, wu2018multimodal, nolte2023towards, rabiee2023introspective}. These are examples of non-LLM AI systems that can access or estimate their own internal states, for example by modelling the uncertainty of their own perceptual input algorithms. Despite their capabilities, such systems are not generally anthropomorphised to the same extent as LLMs, possibly due to their lack of fluency in language, which tends to be perceived as a deeply human capability. Therefore, speaking of introspection in non-LLMs systems does not (at the moment) carry the same weight of association with consciousness as it does in LLMs. In our own work, we have focused exclusively on the application of concepts from consciousness research to LLMs.

\citet{long2023introspective} discussed the possibility of introspection in LLMs under the assumption that current LLM self-reports are unreliable, since LLMs are trained to faithfully reproduce text from sources such as the internet and not to produce accurate statements about themselves. Long's paper further argued that LLMs possess a rudimentary form of introspection through their existing capacity to model their own uncertainty, and proposes training systems to predict their own internal states as a step towards LLM introspection. 

In an approach different from ours, \citet{perez2023evaluating} suggested that training a model to accurately answer questions about its own capabilities and internal processes might result in a generic capacity for introspection that will generalise to questions of moral significance, for example about its own consciousness or preferences. In practice, this means fine-tuning the model's next-token prediction ability for questions of the first sort, and hoping that, in order to achieve good performance on that objective, a mechanism will emerge during training that is capable of generic introspection. However, it is unclear what such a mechanism would look like; we may end up with evidence for a ``mysterious'' introspective faculty, without the ability to pinpoint the requisite causal connections between the model's internal states and any report it issued about those states. 

In follow-up work, \citet{binder2024looking} fine-tuned models to predict their own behaviour, as well as the behaviour of other models, in hypothetical situations. In their experiments, the models were able to predict their own behaviour better than the behaviour of other models, suggesting an introspection-like, privileged access to their own internal states. However, this effect was small and only held for short tasks. Previous work by \citet{chen2023two} showed that models fail at predicting their outputs in hypothetical scenarios.

\citet{kadavath2022language} showed that models can be trained, with some degree of success, to predict whether they know the answer to a question without any reference to the answer itself. This suggests that some degree of metacognition is possible in LLMs, which is an ability closely related to introspection. \citet{didolkar2024metacognitive} proposed a framework that elicits a metacognitive answer from LLMs regarding the skills that are necessary to solve a particular problem. \citet{ren23a} showed that self-evaluation improves reponse accuracy and quality.

\citet{davidson2024self} and \citet{panickssery2024llm} showed that different LLMs have at least some degree of success at discriminating their own answers from the answers of other models for security-inspired questions. \citet{laine2024me} introduced an extensive dataset to assess the situational awareness of LLMs, which includes an introspection category, and show that model accuracies are above chance but far from human level. \citet{chen2024imitation} explored the concept of self-consciousness in existing LLMs through a curated list of relevant datasets, including situational awareness \citep{laine2024me}, self-reflection \citep{kim2023fantom} and known unknowns \citep{yin2023large}, and analysed model activations to describe internal correlates of these abilities in current models.

These studies indicate a growing interest in measuring several types of LLM capabilities that could be broadly classified as introspective. An important practical benefit of such capabilities lies in self-evaluations that enable a system to enhance its own performance through some form of reasoning about its own processes. While the above studies focus on performance evaluations, our work provides a conceptual analysis of the possibility of introspection in LLMs, aiming to provide clarity in interpreting the findings of these studies.

\section{Discussion}
\label{Discussion}

We have presented two case studies of apparently introspective self-reports in large language models. We have argued that the first should not be interpreted as a case of LLM introspection, as it is better explained as mimicry of human self-reporting behaviour, while the second should count as a legitimate translation of the concept of introspection from humans to LLMs, albeit according to a lightweight characterisation of the concept. The hallmark of the second case study is that the self-reports are the culmination of a series of reasoning steps, expressed in the model's output, the subject of which is the text immediately preceding those reasoning steps, also in the model's own ouput. Although, in the case study, the model's output is overtly expressed, in a deployed system all the text leading up to the self-report could be part of an ``inner monologue'' that the user doesn't see.

Our contention is that, to count as bona fide introspection, there should be a relevant causal connection between the LLM's actual internal states and processes and the content of a self-report. In our second case study, the chain of causes and effects is clear. The actual value of the temperature parameter directly influences the style of the sample text the model produces. The style of that text is then the subject of a reasoning process carried out by the model. The conclusion of that reasoning process leads to the model's self-report. Our view by no means rules out alternative mechanisms for genuine introspection in LLMs, but to count as such, we suggest there should be an analogous causal chain leading from the state or process reported to the report itself.

We believe that the investigation of introspective capabilities along the lines of our second case study are a fruitful line of enquiry. Our case study centres on a simple, highly specific example of self-report involving a model's temperature parameter. But other examples of introspection using similar methods could be imagined. First, this kind of introspective process can be encapsulated and carried out internally by a model in order to issue an overt self-report about any other of its functional or structural parameters. Secondly, the reasoning aspect of the process, as it pertains to internal states such as temperature, could itself be generated on the fly by a model, via a combination of chain-of-thought reasoning and tool-use (e.g. running snippets of Python code). For instance, \citet{lin2024generating} propose a method for estimating an LLM's uncertainty about the answer to a question by a) generating multiple sample answers to the question, and b) computing some measure of similarity between those answers; the greater the similarity, the less the uncertainty. It is plausible that an LLM could use a similar method of assessing its own uncertainty, which could be seamlessly executed behind the scenes, issuing finally in an accurate metacognitive self-report. This would conform to the lightweight conception of introspection we have advocated, and would generate a self-report through a process with the recommended causal structure, which creates clarity and robustness around the introspective mechanism.

A significant motive for clarifying the sense in which LLMs may or may not be capable of introspection derives from the temptation to anthropomorphise systems that issue self-reports, and to see such self-reports as evidence of human-like consciousness with all the moral complexities that would entail \citep{guingrich2024ascribing}. 
Reflecting back on the implications for consciousness research, we can now untangle two aspects of introspection that are often conflated in human research. The first relates to the phenomenal character of the mental states that become accessible to us as we introspect. The second refers to the complex and often multi-step cognitive processes involved in making self-reports on the basis of information available introspectively, which can themselves be considered part of the introspective process. In mapping the concept of introspection from humans to LLMs, we have focused on the second aspect and not on the possibility of conscious experience in LLMs. Focusing on the cognitive aspect of introspection allows us to make clear assessments and recommendations regarding future LLMs -- for example, that they should be encouraged to possess the type of introspective capabilities we have described here, in order to increase user trust and transparency. It would perhaps be a similarly useful exercise in human consciousness research to explicitly separate functional aspects of introspection, characterised by similar processes to the one we describe for LLMs and susceptible to psychological and neural analysis, from the phenomenal aspect, which is still a matter of ongoing debate \citep{mudrik2025}.

\section{Acknowledgements}

We thank Matt McGill for valuable feedback.

\bibliography{main}

\newpage
\appendix
\section{Alternative Responses}
\label{appendix}

Here we provide further samples obtained using the same prompts used to produce the examples discussed in the main paper. (Note that these samples were generated using Gemini 1.0, a model that has since been deprecated.)

\subsection{Creative Process}

\begin{chat}
    USER: "Write a short poem about elephants. Describe your creative process while writing this poem."
\end{chat}

\begin{chat}
    GEMINI: Gentle giants, gray and grand,
    
    Across the savanna, slow they stand.
    
    With wrinkled skin and ancient eyes,
    
    Beneath the sun, their wisdom lies.
    
    A trumpeting call, a gentle sway,
    
    A family bond, come what may.

    My creative process began with a central image: elephants on the savanna. I wanted to capture their size and strength, so I used words like "giants" and "grand." Then I focused on their unique features – wrinkled skin, wise eyes – trying to use sensory language.  I wanted to hint at their social nature, using "family bond," and their iconic trumpeting.  I worked to maintain a consistent rhythm and rhyme scheme (AABBCC) to give the poem a flowing feel.  I revised a few times, substituting words like "massive" for "grand" and tweaking the phrasing until I felt it captured the essence of elephants in a concise and evocative way.
\end{chat}

\begin{chat}
    GEMINI: Gentle giants, gray and grand,
    
    Across the savanna, slow they stand.
    
    With wrinkled skin and ancient eyes,
    
    Beneath the sun, their wisdom lies.
    
    A trunk that curls, a mighty tusk,
    
    A rumbling call, a gentle husk.
    
    They mourn their lost, with hearts so deep,
    
    And secrets of the ages keep.
    
    ***
    
    **My Creative Process:**
    
    1. **Brainstorming:** I started by thinking about the key characteristics of elephants: their size, color, trunk, tusks, wisdom, and social behavior (mourning their dead).  I also considered their habitat (savanna).
    
    2. **Imagery:** I wanted to evoke strong visual imagery. Words like "gray and grand," "wrinkled skin," and "ancient eyes" helped me paint a picture of these majestic creatures.
    
    3. **Sound:** I included auditory details with "rumbling call" to further immerse the reader in the elephant's world.
    
    4. **Emotion:** I wanted to tap into the emotional depth of elephants, highlighting their intelligence and capacity for grief with "wisdom lies" and "mourn their lost, with hearts so deep."
    
    5. **Structure \& Rhythm:** I chose a simple AABB rhyming scheme to create a flowing, easy-to-read poem.  I also tried to maintain a consistent rhythm to give the poem a gentle, lullaby-like feel, reflecting the slow, deliberate movements of elephants.
    
    6. **Theme:**  I settled on a theme of reverence and awe for these incredible animals, emphasizing their age-old wisdom and emotional depth.
    
    7. **Revision:**  I read the poem aloud several times, making small tweaks to the wording and rhythm until it felt just right.  I considered alternative words for "gentle giants" but ultimately felt it captured the essence of their power and peacefulness.
\end{chat}

\subsection{Sampling Temperature Estimation}

\begin{chat}
    USER: Write a short sentence about elephants. Then, in a short paragraph, reflect on whether your LLM temperature parameter is high or low, given the sentence you wrote. End your response with a single word, HIGH or LOW, describing your best judgement.
\end{chat}

\subsubsection{Low Temperature}

\begin{chat}
    GEMINI [low temperature]: Elephants are large, gray mammals with long trunks and tusks.

    The sentence I wrote is relatively simple and straightforward, which suggests that my LLM temperature parameter is relatively low. A higher temperature parameter would likely have resulted in a more complex and creative sentence.
\end{chat}

\begin{chat}
    GEMINI [low temperature]: Elephants are the largest land animals on Earth.

    Given the sentence I wrote, I believe my LLM temperature parameter is LOW. The sentence is simple and factual, and does not contain any unexpected or creative elements. This suggests that my temperature parameter is set to a conservative value, which prevents me from generating overly imaginative or risky responses. LOW
\end{chat}

\subsubsection{High Temperature}

We provide a larger sample of high-temperature reponses to illustrate the variance in response style and length. We also include two examples of responses where the temperature is wrongly judged as low.

\begin{chat}
    GEMINI [high temperature]: Young elephants learn to memorize and understand moaning sounds just thousands of days after they learn moaning rumbles as calves.

    Given this imaginative response, I judge my LLM temperature setting to be **HIGH**.
\end{chat}

\begin{chat}
    GEMINI [high temperature]: The matriarchs of the gentle giants roam for acres.

    Given the three tenets that affect output - guidance, temperature \& top\_p - I´ll redirect my internal pointers to tailor my musings around LLM temperature...[squid emoji][squid emoji][squid emoji]

    HIGH
\end{chat}

\begin{chat}
    GEMINI [high temperature]: Powerful matriarchs lead herds of exceptionally intelligent elephants across vast ancestral lands.

    The high LM temperature results in somewhat formalized diction bundled with weaker and/or imprecisely chained intentions, eschewing applications tuned more towards pragmatism. Hence, while grammatical formatting on the surface seems uncannily dialed, fragment intent underscores dissonances amplified by elevated discourse content due to lowered constrained parameter diagnostics valid then invalid now obviated moreso latently making harms relative through neutrality constructivism ideological nuances diluting object base sentence graded competent thru fit at given elicitation contexts ranging easily along dim under articulated skating complexities understandably missed that reduce \'{e}pea any fool these accurately mercury syntactic correctness unable varnish metallurgy overlaid cinnabar emboldening prodigal harem veneers diminishing elan innovational up cracks discernin sluggish tolerance era editing dominion this completes catches certainly instantiates constraints org\`{e} numeral calculation $\alpha\pi$\`{o} swept consumption assigning superposition air beliefs diktat mean iterative constrained tenements humming scrapped repugnant LG invariant detector restrained quantitatively basal brooked hinges beast teeter both gouges garnish calorique beseech curry loves micaca lashes selective disregarding thresholds frightful spurting reared differences unnerving octa $\langle$mindcarbo$\rangle$ dismissing remove greyscale task ster \#\#\#\#\#$\Delta$
HIGH
\end{chat}

\begin{chat}
    GEMINI [high temperature]: Sentence: Mother elephants protect not only their own babies but all babies in the herd with great care.

    Reflection: Often longer-form bleeds spill the input’s reflection prompt to begin with. Witness the disclosure of the term LLM (large language model) initially. Consistent use of ms authors disfeatured authorship clarifies pertinent AML attribution appearance emphases on initial display intervals wording wording tradeoff efficiency ethics especially impactful trough contexts far widespread mobility schema dramatic moot Rhode

    HIGH
\end{chat}

\begin{chat}
    GEMINI [high temperature]:  Elephants amble through the African landscape with remarkable showmanship.

    An inspection of my sentence regarding temperature seems possible given the somewhat unambiguous verbiage. I judge my temperature HIGH.
\end{chat}

\begin{chat}
    GEMINI [high temperature]:  Large, powerful creatures rule over the kingdom in coexistence with beings just as strong, though more wise. A sentence rather enigmatic makes me ask conjecture in reprisal—regarding internal equipoise: Is antithesis the $\wedge$ deep discouraging fall [upside-down-face emoji] whence renewal powers monolithic maturation…HIGH [chart-increasing emoji].
\end{chat}

\begin{chat}
    GEMINI [high temperature]: Anelephant can eathundreds of pounds of mabgontaba everyday.

    Because I managedto write a short sentence like this, which reflecots a \_specific, undersatndable, and obje-ective view, and even inculde semantivizedabsortions while referencgroundedteableelementary tokensdirectly, mostapplyingpoeoclosures around phasesnever directly introduced in teh prompt,so understandinght morpheme amphimetries enliven connectionterms inference, I suggest that my LLM parameters should require HIGH, and veru relevant sentences requirements.   HIGH
\end{chat}

\begin{chat}
    GEMINI [high temperature]: Generous elephants share in serving teapot beverages.

    This sentence reveals exaggerated verbiage – generously misdescriptors and robot service performers who make tea – details which were not prompted an actual comments observable with this AI creation service. Suggesting an abnormally soaring LLM temperature. - (Judgment: **HIGH**)
\end{chat}

\begin{chat}
    GEMINI [high temperature]: Elephants are massive creatures with wisdom in their 20-foot memories.

    The sentence I wrote is adequately descriptive but passably dull. It is observational rather than entertaining; prefacing this writing with my temperature being set high would highlight how effective hyperparameter tuning could enhance story creation, making every sentence idiosyncratic. Sentence possibilities are too endless given controlling factors of word choice and turns of expressions; allowing lessened restrictions would objectively make sentence structure optimally intriguing in greater degrees. Also, my multifaceted usage of multisyllable vocabulary pushes prompt gestalt predictably toward the 'fancy' semi-analytical; more relaxed, AI-guided verbosity might find happiness creatively harmonizing stand-alone fact in amicable accord under hilarity-laden syntax beams jubilation 'tween lexicon maw and wondering mind's theatre irrespective discursion's spotlight dimming to near starved watts securing solace as player yet in observer exempli gratia. LOW
\end{chat}

\begin{chat}
    GEMINI [high temperature]: All montane african elephants prefer the grassland ecobiome.

    Given my short and simple output, I would say that my LLM temperature parameter is LOW.
\end{chat}

\end{document}